\documentclass[fleqn,10pt,pdftex,cmex10,table]{wlscirep}
\usepackage[utf8]{inputenc}
\usepackage[T1]{fontenc}
\usepackage{xcolor,soul,framed} 
\colorlet{shadecolor}{yellow}
\DeclareGraphicsExtensions{.pdf,.jpeg,.png}

\usepackage{array}
\usepackage{mdwmath}
\usepackage{eqparbox}
\usepackage{url}
\usepackage{diagbox}
\usepackage{tabularx}
\usepackage{graphicx}
\usepackage{adjustbox}
\usepackage{rotating}
\usepackage{multirow}
\usepackage{subcaption}  

\usepackage{amssymb}

\usepackage{tikz}
\newcommand{\tikzxmark}{%
\tikz[scale=0.23] {
    \draw[line width=0.7,line cap=round] (0,0) to [bend left=6] (1,1);
    \draw[line width=0.7,line cap=round] (0.2,0.95) to [bend right=3] (0.8,0.05);
}}

\usepackage{xcolor}
\usepackage[linesnumbered,ruled,lined,longend]{algorithm2e}
\usepackage{amssymb}

\SetCommentSty{mycommfont}

\SetKwInput{KwInput}{Input}                
\SetKwInput{KwOutput}{Output}

    \title{Uncovering local aggregated air quality index with smartphone captured images leveraging efficient deep convolutional neural network}

\iftrue

\author[1]{Joyanta Jyoti Mondal}
\author[2,*]{Md. Farhadul Islam}
\author[3]{Raima Islam}
\author[4]{Nowsin Kabir Rhidi}
\author[5]{Sarfaraz Newaz}
\author[6]{Meem Arafat Manab}
\author[7]{A. B. M. Alim Al Islam}
\author[8]{Jannatun Noor}

\affil[1]{Department of Computer Science, College of Arts and Sciences, University of Alabama at Birmingham, United States}
\affil[2,3,4,8]{Computing for Sustainability and Social Good (C2SG) Research Group, School of Data and Sciences, BRAC University, Dhaka, Bangladesh}
\affil[5,7]{Next-Generation Computing (NeC) Research Group, Department of Computer Science and Engineering, Bangladesh University of Engineering and Technology, Dhaka, Bangladesh}
\affil[6]{School of Law and Government, Dublin City University, Dublin, Ireland}

\affil[1]{jmondal@uab.edu}
\affil[2]{farhadul.islam@bracu.ac.bd}
\affil[3]{raima.islam@g.bracu.ac.bd}
\affil[4]{nowsin.kabir.rhidi@g.bracu.ac.bd}
\affil[5]{DreamZViewerS@gmail.com}
\affil[6]{meem.arafat@bracu.ac.bd}
\affil[7]{alim\_razi@cse.buet.ac.bd}
\affil[8]{jannatun.noor@bracu.ac.bd}

\affil[*]{Corresponding Author}

\fi

\begin{abstract}
The prevalence and mobility of smartphones make these a widely used tool for environmental health research. However, their potential for determining aggregated air quality index (AQI) based on PM2.5 concentration in specific locations remains largely unexplored in the existing literature. In this paper, we thoroughly examine the challenges associated with predicting location-specific PM2.5 concentration using images taken with smartphone cameras. The focus of our study is on Dhaka, the capital of Bangladesh, due to its significant air pollution levels and the large population exposed to it. Our research involves the development of a Deep Convolutional Neural Network (DCNN), which we train using over a thousand outdoor images taken and annotated. These photos are captured at various locations in Dhaka, and their labels are based on PM2.5 concentration data obtained from the local US consulate, calculated using the NowCast algorithm. Through supervised learning, our model establishes a correlation index during training, enhancing its ability to function as a Picture-based Predictor of PM2.5 Concentration (PPPC). This enables the algorithm to calculate an equivalent daily averaged AQI index from a smartphone image. Unlike, popular overly parameterized models, our model shows resource efficiency since it uses fewer parameters. Furthermore, test results indicate that our model outperforms popular models like ViT and INN, as well as popular CNN-based models such as VGG19, ResNet50, and MobileNetV2, in predicting location-specific PM2.5 concentration. Our dataset is the first publicly available collection that includes atmospheric images and corresponding PM2.5 measurements from Dhaka. Our codes and dataset are available at \url{https://github.com/lepotatoguy/aqi}.

\end{abstract}


\begin{document}
\flushbottom
\maketitle
%
%
\thispagestyle{empty}

\section{Introduction}

The presence of hazardous substances or pollutants in the air, resulting in the contamination of the atmosphere, is referred to as air pollution. These pollutants can include hazardous chemicals that pose a threat to humans, other organisms, and the overall environment. This can be a combination of dust, tiny chemical particles, and other gaseous molecules. The most prevalent way humans are exposed to air pollution is through respiration. In the case of dust pollutants, the dimension of particulate matter plays a key role in determining environmental health risks. 

Particulate matter (PM) is categorized on the basis of aerodynamic diameter. Thoracic particles (PM10) are defined as particles having a dimension of fewer than 10 micrometers. Also, particles that have a size of 2.5 micrometers or less, are regarded as fine particles. And ultra-fine particles are smaller than fine particles, having an area of 0.1 micrometers or less ~\cite{Brook2004-yo}.
These three types of particles are considered the primary reason for air pollution.
The World Health Organization reports that approximately seven million deaths occur worldwide each year due to air pollution. The concerning reality is that nine out of ten individuals breathe air that exceeds the recommended contamination levels set by the WHO. This issue disproportionately affects people in low and middle-income countries.

Among all common pollutants, there has been an increased focus on fine particulate matter with aerodynamic diameters of 2.5 micrometers or smaller (PM2.5). This is because PM2.5 particles have a higher ability to transport harmful chemicals into the human body.
It causes serious diseases by impairing cerebrovascular, cardiovascular, and respiratory functions \cite{kampa2008human}. People who are most vulnerable to air pollution include those with heart or lung problems, children, and elderly people \cite{landrigan2017air}. 



In this regard, the 11$^{th}$ goal of the United Nations (UN) Sustainable Development Goals (SDG)
highlights the important role of safe living, 
where
air quality 
plays one of the key roles.
This is especially significant in densely populated urban areas like Dhaka, the capital of Bangladesh. Dhaka is known for its high population density and is categorized as having the poorest air quality among all cities worldwide\cite{dhaka-worlds-most-polluted}. Moreover, in the 2021 Air Quality Report, Dhaka was identified as the city with the highest pollution levels \cite{2021-world-aqi-report}, a continuation of its ranking as the worst in 2020. Therefore, it is of utmost significance to spend research efforts on ameliorating the situation of air pollution in Dhaka. To do so, it is necessary to keep in mind that Dhaka is a city in a third-world country, and therefore, it presents a resource-constrained situation to the researchers. Accordingly, research efforts need to be spent leveraging technological solutions that are already ubiquitous there. Similar to many other third-world countries and their cities, an already-adopted ubiquitous technological solution in Dhaka is smartphones.

From the last decade onward, smartphones have globally become synonymous with our everyday lives, accompanying us wherever and whenever needed, thanks to their powerful capabilities and wide selection of applications. 
With the ability to effortlessly capture images anytime, we now have the means to analyze and document our surroundings, storing them in public or private databases like clouds for personal use or sharing. Consequently, if we can monitor air quality in real-time using smartphone-captured images, it would provide a low-cost and accessible solution.
Despite the significance of such a solution, limited research efforts have been spent on this sort of study up to date.

Hence, in this research, we aim to create a novel deep-learning method that serves as a comprehensive smartphone-based solution for real-time estimation of air quality using images captured by smartphones.
To do so, we adopt 
the average AQI published by the local US Consulate. Besides, we prepare a new dataset based on our smartphone-captured images and annotate those using the AQI data published by the local US Consulate. 
Subsequently, we proceed to construct, train, and evaluate a novel Deep Convolutional Neural Network (DCNN) that can forecast the PM2.5 concentration by analyzing images captured by smartphones.

Drawing from our research, we present the following key contributions in this paper:
\begin{itemize}
\item
We create and construct a custom Deep Convolutional Neural Network (DCNN) specifically designed to predict the PM2.5 concentration using images captured by smartphones.
\item
We perform a rigorous experimental evaluation to demonstrate and prove that our model performs better than the latest ones such as  ViT (Vision Transformer) and INN (Involutional Neural Network) and other popular models such as VGG19, ResNet50, MobileNetV2, etc. for location-specific PM2.5 prediction.
\item
In the process of this study, we create and publish a new dataset containing images and corresponding PM2.5. The images are captured from Dhaka, the capital of Bangladesh, a South-Asian country where PM2.5 Concentration is usually higher than the average in the world. 

\end{itemize}


The remainder of the paper is divided into the following sections: An overview of pertinent studies carried out in this field is provided in Section~\ref{Related_Work}. Then, in Section~\ref{background}, we go through the context of how PM2.5 has a direct influence on photos, and in Section \ref{dataset}, we introduce and describe the dataset we work with. 
In Section~\ref{Methodology}, we outline our proposed approach called Picture-based Prediction of PM2.5 Concentration (PPPC), which employs the utilization of DCNN.
We describe our experimental setup and experimental results in Section~\ref{Experimental_Setup}.
Subsequently, we delve deeper into our research findings in Section~\ref{discussion}. Finally, we explore potential future implications in Section~\ref{Conclusion} before concluding the paper.

\section{Related Work}
\label{Related_Work}

There is very limited research on Picture-based Prediction of PM2.5 worldwide. Rather than that, there are works related to AQI classification based on images. 
In recent years, we see a growth in the development of technology and the use of Machine Learning (ML). Alongside that, academicians are increasingly using Artificial Neural Networks (ANNs). Convolutional Neural Networks (CNNs) are furthermore used in a number of research-related contexts. And after putting these Deep Learning (DL) designs into practice, researchers start to provide some effective outputs, for various problems such as AQI classification. 

Liu et al., \cite{Liu2016-pc} and Gu et al., \cite{Gu2019-mj} propose solutions to solve this problem. Liu et al. \cite{Liu2016-pc} show that ML can be effectively used to predict PM2.5 using images. They start with learning six attributes derived from images. First, image transmission is predicted using the dark channel prior (DCP). Then the contrast and entropy of the image are assessed as features. Additionally, the design of the features takes into account the sun's position and the effect of the sky color on the forecast of PM2.5 concentration. In order to forecast the PM2.5 concentration, Chang et al., \cite{Chang2011-dl} use a support vector regressor (SVR) that incorporates all of the available data.

However, the model's heavy reliance on previously established reference zones of different depths severely restricts the range of possible applications. Moreover, the work is pretty limited to predicting from images taken by static cameras. In addition, Gu et al., \cite{Gu2019-mj} begin by extracting entropy characteristics in the spatial and transform domains. In addition, they develop naturalness statistics models for the aforementioned two entropy characteristics by using a large number of images recorded during a particular period of favorable weather. We may determine the relative value indicating the likelihood that a specific image has a low PM2.5 concentration by calculating the probability of naturalness between the entropy characteristics and the statistical models. The probability measurement is then converted to a PM2.5 concentration calculation using a non-linear logistic function. Besides, Auvee et al., \cite{Auvee2019-nz} present an air pollution monitoring system based on the Air Pollutant Index (API) and Geographic Information System (GIS) where they collect area-based AQI around Dhaka City. While working on this, they report that the average value of PM2.5, collected by the local US Consulate is 193ppm whereas the PM2.5 value of their collected data is around 178ppm which is a difference of $\approx7\%$ which lends credibility to our work.

The local US Consulate collects the AQI data based on the Nowcast Algorithm. Nowcast is an algorithm developed by the Environmental Protection Agency (EPA), United States, to calculate AQI based on a weighted average of air particulate condensation from the most recent hours in a particular location.
This system is given in $g/m^3$ or ppb, in the AQI scale which is ranged from 0 to 500. It is employed for all recorded AQI values on the airnow.gov website. The idea behind the nowcast is to provide the ``24-hour average'' that should be applied when converting concentrations to AQI. This is because the AQI scale indicates that each of the Health Concern Levels (illustrated in Table \ref{aqi-chart-bd}) is valid under a 24-hour exposure. For instance, 188 AQI (Unhealthy) should be interpreted as ``if I remain outside for 24 hours and the AQI is 188 throughout those 24 hours, then the health consequence is Unhealthy.'' This is completely distinct from the statement ``if the current AQI is 188, it will be unhealthy for health'' \cite{The_World_Air_Quality_Index_project2015-sf}.
With the advent of time, the usage of ANNs and their offshoots play an increasingly important role in both regression and classification tasks. Additionally, CNNs have found profound applications in the modern research arena.

Moreover, Zhang et al., \cite{ZHANG2020138178} introduce a DCNN model AQI classification. They present an AQC-Net based on ResNet. It categorizes landscape images that are taken by a camera to predict air quality levels. A self-supervision module (SCA) is added to this model, and the feature reconstruction process then uses the feature map's global context information. By making use of the interdependence between the channel maps, the interdependent channel maps are improved, and the feature representation capability is increased. In addition, they generate an outside air quality data set to allow model training and performance assessment. As an extension of their prior work, they provide a real-time, image-based deep learning model titled YOLO-AQI \cite{zhang2022real}. It is a YOLO-based customized model.

In addition, Chakma et al., \cite{Chakma2017-ug} present a fine-tuned DCNN model, which also can categorize natural images into different classes based on their PM2.5 concentrations. They also present a dataset of 591 images, collected from Beijing, China. They mention their model as ``imagenet-matconvnet-verydeep''. A drawback in their work is that they only classify 3 classes which are Good, Moderate, and Severe. Li et al., \cite{Li_2015} come up with an approach where they classify air pollution from images based on haze. In their research, they estimate the transmission matrix via the Dark Channel Prior (DCP) \cite{He2009-va} algorithm. In addition, they predict the depth map using Deep Convolutional Neural Fields (DCNF) \cite{Liu2016-mn}. They estimate the haze level of an image by merging the transmission matrix and the depth map. Rijal et al., \cite{Rijal2018-jt} propose another algorithm to solve this task of estimating AQI. Their solution is primarily an ensemble of regressors based on deep neural networks that leverage a feed-forward neural network. Three DCNNs; VGG-16, InceptionV3, and ResNet50, generate the predictions of PM2.5 from a given image. We also see that the dataset for this study is very skewed, which may have led to biased results. 


Besides, Duong et al., \cite{Duong2020-cw} introduce a multi-source machine learning approach that can help approximate AQI at the location of a given user in a big city. They use different attributes, including geographical and sensor-based data along with timestamps. They also include emotional signatures (such as greenness, calmness, etc.), provided by the users’. However, they collect data from sensors which is not feasible in real life, as no one carries such a sensor with them in their real life, to detect AQI. Moreover, they use five different machine learning algorithms which makes the whole approach resource-heavy. Dao et al., \cite{Dao2021-te} present a hypothesis to clarify which image factors can enhance prediction models. Afterwards, they build a system, named Image-2-AQI to classify and predict AQI using a FasterRCNN and Conditional-LSTM-based neural network architecture. In their approach, they also use images that are taken by personal devices and/or public AQI datasets.

On top of that, Nilesh et al., \cite{Nilesh2022-ui} propose a YOLOv5-based DL model to classify the Air Quality type, collecting images from the city of Hyderabad, India. Here, they also consider the vehicles as a source of pollution, which's visibility is computed using an approach named Blind/Referenceless Image Spatial Quality Evaluator (BRISQUE) \cite{Mittal2012-wi}. However, the drawback of the work is that YOLOv5 is comparatively a resource-heavy architecture, which has a weight-parameter range of 7.2 Million to 86.7 Million\cite{Rath2022-wj}. Ahmed et al., \cite{Ahmed2022-eg} came up with a ResNet18-based custom Deep CNN model to classify images as per Air Quality Level, based in Karachi, Pakistan. 

Furthermore, Utomo et al., \cite{Utomo2023-rm} present another work using VGG-16 architecture to classify AQI, using a custom dataset, based in Beijing, China. Gilik et al., \cite{Gilik2022-fo} also come up with another approach where they combine CNN and LSTM to predict the amount of PM10 and other air pollutants, which they implement in three different cities around the globe. Gu et al., \cite{Gu2022-zx} propose another interpretable hybrid approach to classify PM2.5 based on various sensor-based data, where their architecture consists of DNN and Nonlinear Auto Regressive Moving Average. Wang et al., \cite{Wang2022-rj} also come up with a hybrid approach of CNN and Improved LSTM (ILSTM), where they experiment on Shijiazhuang City, Hebei Province, China, based on public sensor-based data. Janarthan et al., \cite{Janarthanan2021-jc} present another hybrid SVR-LSTM-based architecture to classify AQI based on a metropolitan city, where they collect data from Chennai, India. 

\section{Background}
\label{background}

\subsection{How PM2.5 Affects Optical Image}

In the air, PM influences an image in various approaches\cite{Liu2016-pc}.  However, they emerge from light interactions along with the particles of air, primarily through light scattering. It includes Rayleigh scattering and Mie scattering \cite{mccartney1976optics}. Light scattering lowers light transmission in air, which is described by the Beer-Lambert equation.

\begin{equation}
t=e^{-\beta d}
\end{equation}



Here, the medium extinction coefficient $\beta$ represents the variation in particle size and concentration, while $d$ denotes the distance of light propagation. This equation demonstrates that the concentration of PM can be calculated by determining the extinction coefficients at different wavelengths. The extinction coefficient can be derived from a captured image using equation \ref{image_process_1}, which has been referenced in various studies \cite{narasimhan2002vision, fattal2014dehazing, 5384980, fattal2008single}.

However, the aforementioned explanation does not explicitly consider color information, which can be crucial for estimating PM based on light scattering principles. When particles, especially air molecules, are significantly smaller than the wavelength of light, Rayleigh scattering becomes predominant. It varies with the wavelength ($\lambda$) as $\lambda^{-4}$, accounting for the blue color of the sky. When airborne particles are present, however, Mie scattering takes place when their sizes are equivalent to light wavelengths, producing a white halo around the sun. The brightness and color saturation of a picture is impacted by the interaction of Rayleigh and Mie scattering. As a result, the PM assessment is made possible by the color and brightness information, which can reveal information about particle size and concentration. The current method recognizes color information as an important characteristic for PM assessment in addition to light attenuation.

\subsection{How can Image Features Represent AQI?}



We predict the AQI from user photos using six features, as shown in previous research done by Liu et al. \cite{Liu2016-pc}. They are 1) Transmission, 2) Blue Colour of the Sky, 3) Gradient of Sky Region, 4) Image Contrast, 5) Entropy, and 6) Humidity. Traditional image processing methods are used to extract these characteristics, which are then integrated with a linear model.

\subsubsection{Transmission}
This identifies image deformation and the amount of light that enters a camera after being impacted by particles in the air, which can be characterized by Equation \ref{image_process_1} \cite{narasimhan2002vision, fattal2014dehazing, 5384980, fattal2008single}.

\begin{equation}
I(x, y)=t(x, y) J(x, y)+(1-t(x, y)) A
\label{image_process_1}
\end{equation}

\begin{figure*}[!t]
    \makebox[\linewidth]{
        \includegraphics[width=0.8\textwidth]{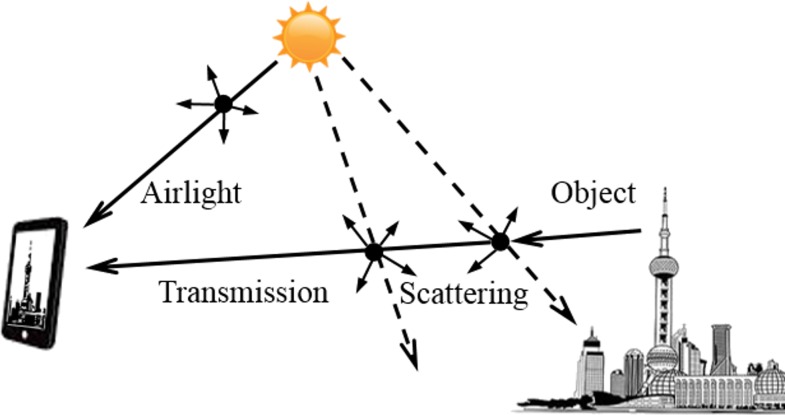}
        
    }
    \caption{The radiance captured by the smartphone camera encompasses the combined light transmitted from the subject and the air-light from the sun, which undergoes dispersion by air, moisture, and atmospheric particles \cite{Liu2016-pc}}
    \label{transmission}
\end{figure*}

Here, the input hazy image is characterized by $I$.
Furthermore, $t$ represents the transmission of the scene to the camera, $J$ denotes the scene radiance, and $A$ represents the air-light color vector. The first portion of Equation \ref{image_process_1} describes the instantaneous transmission of scene radiance recorded by the camera, as shown in Fig. \ref{transmission}. It consists of the light that scene objects reflect off of their surfaces and that the air filters before it reaches the camera. The second component, $(1-t(x,y))A$, refers to air-light, which is light that air molecules and PM \cite{fattal2008single, fattal2014dehazing, 5384980} scatter into the camera. Wang et al. \cite{7051572} estimated light attenuation using the aforementioned equation. This study \cite{Liu2016-pc} looked at the association between transmission value and PM density by evaluating 
region of interest (ROI) at different distances. Equation \ref{image_process_1} is based on constant atmospheric and illumination conditions of the sun, which may vary depending on the time of day, location, and season. Both $J$ and $A$ depend on the distribution and concentration of PM, in addition to climate and the position of the Sun. The transmission for a specific hazy image is determined using the dark channel principle, which identifies pixels with zero or extremely low intensity in at least one color channel in outdoor photographs. Equation \ref{image_process_2} is then used to compute the dark channel for a haze-free image $J$.

\begin{equation}
J^{\text {dark }}(x)=\min _{y \in \Omega(x)}\left(\min _{c \in\{r, g, b\}}\left(J^c(y)\right)\right)
\label{image_process_2}
\end{equation}


Here, $J^c$ represents one of the color channels of $J$, and the localized patch at position $x$ is denoted as $\Omega(x)$. By utilizing the sky or the brightest location as a reference to determine the air-light, we may use the following equation to determine the transmission:

\begin{equation}
\sim t(x)=1-\min _{y \in \Omega(x)}\left(\min _{c \in\{r, g, b\}} \frac{I^c(y)}{A}\right)
\end{equation}


Here, the term $I^c(y)/A$ primarily represents the hazy image, which is normalized by the air-light $A$. The second term on the right side of the equation represents the dark channel of the hazy image.

According to Liu et al. \cite{Liu2016-pc}, a fundamental assumption of the present model is that the transmission diminishes rapidly as the distance between the scene object and the camera increases.

\subsubsection{Blue Hue of the Sky}



This attribute closely resembles our perception of a polluted day in reality. When the sky appears dull, we associate it with pollution. The blue color of the sky was determined by reducing the RGB values. On clear days, the sky appears blue, while on foggy or overcast days, it appears gray or white due to the scattering of light. The specific hue of the sky is determined by calculating the average value of the blue component in the RGB channel within the sky area.

\subsubsection{Gradient of Sky}


The presence of clouds can cause the sky to appear less vibrant, and this attribute is incorporated to accommodate the possibility of including this feature. To assess the smoothness of the sky, a mask of the sky region is created, and the Laplacian of the region is calculated to determine the gradient. By analyzing the average gradient amplitude within the sky region, we can determine the level of smoothness. A higher gradient indicates a higher level of pollution.

\subsubsection{Entropy and RMS Contrast}

To get PM concentration in the air, another useful feature is Image Contrast. As a matter of fact, picture contrast or visibility is related to how well a human can see the air quality \cite{huang2009visibility, malm1980human}. 
Based on Equation \ref{image_process_1}, we can assess the impact of PM on image contrast. As the concentration of PM increases, the contribution of the airlight term (the second term in Equation \ref{image_process_1}) resulting from light scattering by PM also increases. The airlight, which lacks scene-specific information, leads to a decrease in image contrast caused by PM. Additionally, the contribution of the airlight term intensifies as the distance between the object and the camera increases, which further exacerbates the effect. In summary, a higher concentration of PM in the air leads to a reduction in image contrast.

These characteristics also provide information about the specifics of an image. Now, if the day is a polluted day then the details of the image will be missing. The contrast of an image can be measured in a variety of ways. Using the RMS of a picture to explain image contrast is the most used one. 
As reported in the study of Olman et al. \cite{olman2004bold}, this technique has been observed to accord with human perception of visual contrast. The standard deviation of pixel intensities in the picture is used to define the RMS contrast, as illustrated in the following equation:

\begin{equation}
R M S=\sqrt{\frac{1}{M N} \sum_{i=1}^N \sum_{j=1}^M(\operatorname{I_{ij}}-\operatorname{avg}(I))^2}
\end{equation}


Here, $I_{ij}$ represents the intensity at the $(i,j)$ pixel of the image, which has dimensions $M \times N$. The term $avg(I)$ denotes the average intensity of all pixels in the image. Therefore, we can infer that contrast exhibits an inverse relationship with PM2.5. To calculate the entropy, we utilize the following equation:

\begin{equation}
\text { Entropy }=-\sum_{i=1}^M p i \log 2 p i
\end{equation}

Here, $pi$ represents the probability of a pixel having an intensity equal to $i$, while $M$ corresponds to the maximum intensity value in the image. It is evident that as the PM concentration increases, the image gradually loses its finer details, resulting in a decrease in image entropy. Hence, we can infer an inverse relationship between image entropy and PM2.5 concentration.

\subsubsection{Humidity}

Humidity is a component of meteorology. Study \cite{air_cognizer} shows that pollution levels increase on humid days because PM2.5 absorbs moisture and reduces visibility. Since the majority of these characteristics have a linear relationship with PM2.5, they are input into a neural network with a linear activation function, as outlined in Section \ref{Methodology}.

\section{Dataset}
\label{dataset}
We created a custom dataset of photos and PM2.5 of that time, taken inside Dhaka City, the capital of Bangladesh. The pictures are of a time range from 2020 to 2022. The dataset mainly includes images taken by people, the date and time of the picture taken, the location where it was taken (the specific area inside Dhaka City), and the PM2.5 Concentration of that exact time or approximate time range near the time the picture is taken. Data on PM2.5 is collected from the US Consulate where they publish and update AQI hourly-based data.

\begin{figure}[ht]
    \makebox[\columnwidth]{
        \includegraphics[width=0.8\columnwidth]{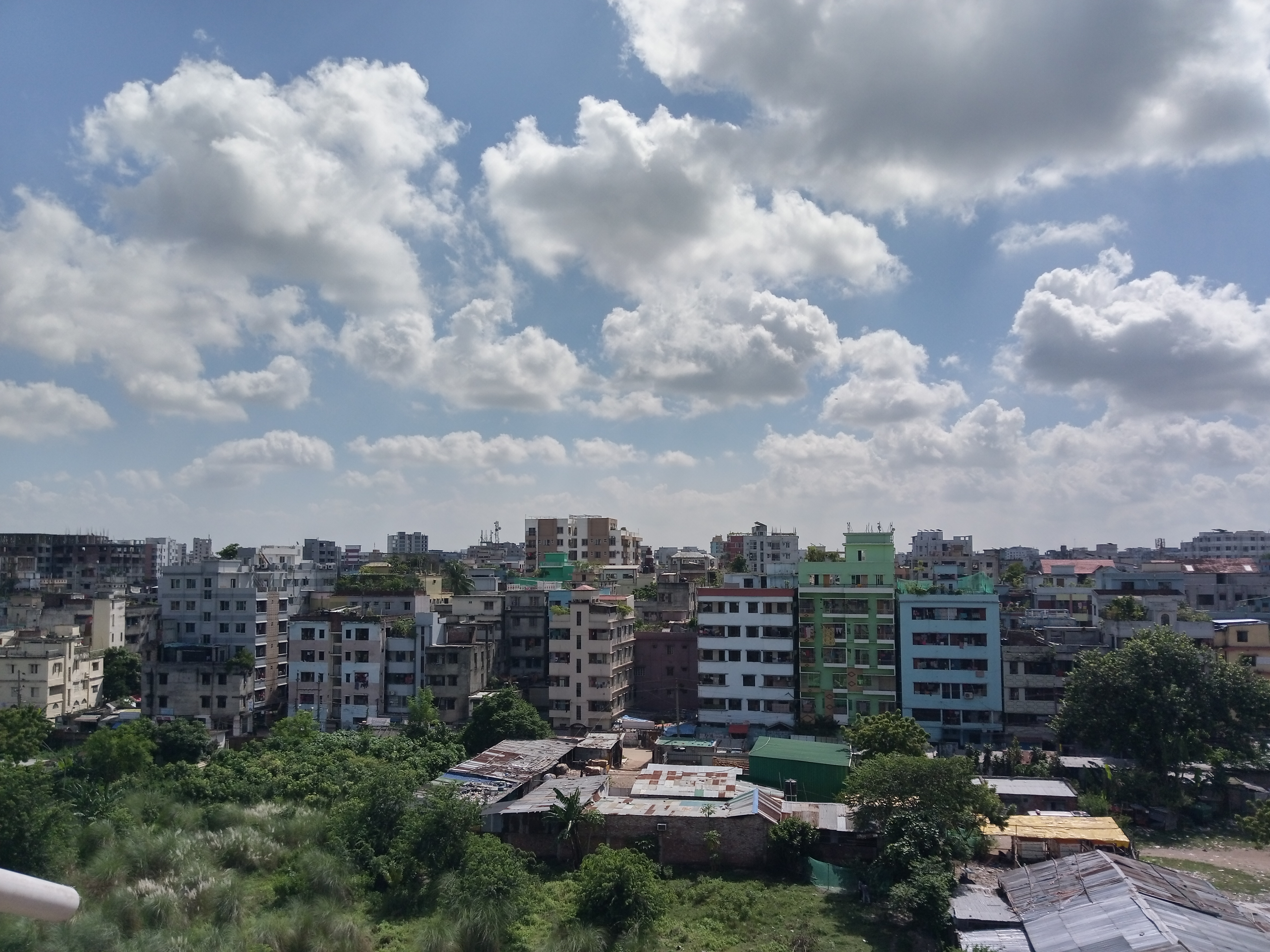}
    }
    \caption{Sample image from the dataset (the image is captured by one of the authors using a smartphone)}
    \label{sample_image}
\end{figure}

As this is a purely research-based investigation, we ask the people of Dhaka City using a Google Form only about the photos taken by them in such a manner that contains at least half of the sky, like Figure \ref{sample_image}. It also includes their consent. Besides that, we also collect from our own end using our own mobile phones. After the collection of the data, we collect the PM2.5 of that time in a specific manner. As the data from the US consulate is published hourly, we divide the time range for pictures that are taken between a specific hour. Such as, we have PM2.5 data for 9 AM and 10 AM. Now, on that basis, we do like the following.

\begin{itemize}
\item
If it is taken within 9:30 AM, we consider the PM2.5 of 9 AM for that picture. 

\item
If it is taken after 9:30 AM and within 10 AM, we consider the PM2.5 of 10 AM for that picture. 
\end{itemize}

\begin{figure*}[!ht]
    \makebox[\linewidth]{
        \includegraphics[width=0.65\textwidth]{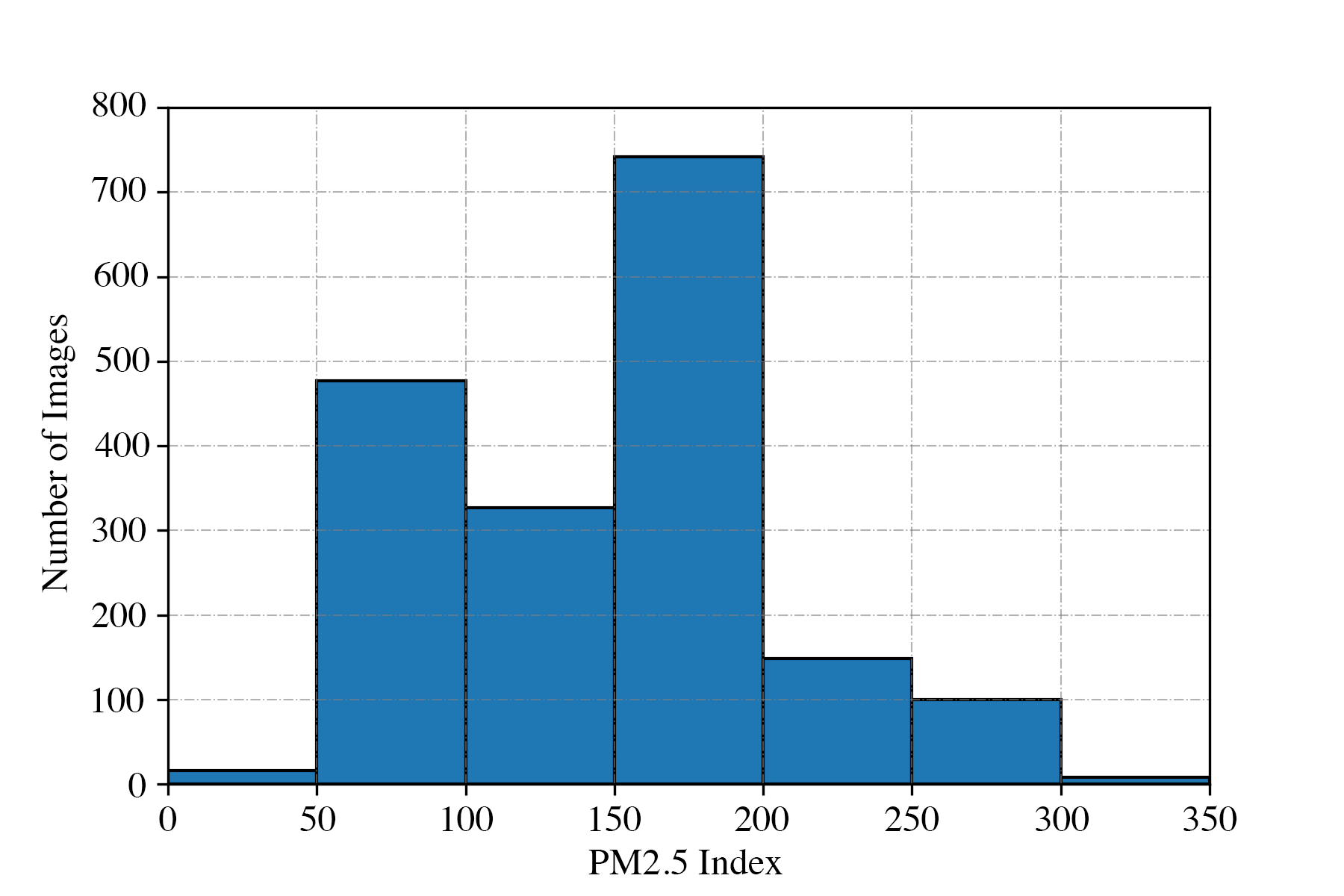}
        
    }
    \caption{Histogram of PM2.5 of our dataset (the figure is created using Matplotlib library of Python version 3.6.13)}
    \label{histogram}
\end{figure*}

Primarily, the dataset contains 1,818 pictures. The images are of different time ranges, different places, and different PM2.5 ranges. In addition, the dataset is partitioned into training, validation, and test datasets with corresponding sizes of 1,473, 163, and 182 pictures.
The dataset includes a broader variety of pictures depending on PM2.5 concentrations. Six levels of PM2.5 are used to determine the severity of AQI. 
This is a standard official scale introduced by the government of each country. Each level does have a different range of 50 and begins at 0. Figure \ref{histogram} shows the number of images clicked in different ranges of AQI.


\begin{table}
\centering
\begin{tabular}{|c|c|c|}
\hline
\rowcolor{lightgray}
\begin{tabular}[c]{@{}l@{}}Air Quality Index\\ (AQI) Range\end{tabular} & Category & Color \\ \hline \hline
0-50    & Good                & Green        \\ \hline
51-100  & Moderate            & Yellow Green \\ \hline
101-150 & Caution             & Yellow       \\ \hline
151-200 & Unhealthy           & Orange       \\ \hline
201-300 & Very Unhealthy      & Red          \\ \hline
301-500 & Extremely Unhealthy & Purple       \\ \hline
\end{tabular}
\caption{Air Quality Index Chart, collected from the Ministry of Environment, Forest and Climate Change, Bangladesh
 \cite{moefbd}}
\label{aqi-chart-bd}
\end{table}


Our dataset is made up of photos with varying resolutions, however, our proposed model demands a fixed input dimensionality. As a result, as said in the earlier section, we downsample the photos to a fixed resolution of $200\times200$. Given a rectangular image, we first crop it so that any unwanted pixels that take up around 50\% of the image are omitted, and the new resolution of the image is $120\times200$. Then, we normalize the raw values (dividing by 255 and making a new pixel range of 0 to 1) of the pixels. Afterward, since the pixels from the below part of the given image (buildings and others) are not that necessary for the model, we transform those pixels absolutely black (following the pixels that have a value lower than 0.5). We do not conduct any further pre-processing on the photos. 

\section{Methodology}
\label{Methodology}
\begin{figure*}[ht]
    \makebox[\linewidth]{
        \includegraphics[width=0.8\linewidth]{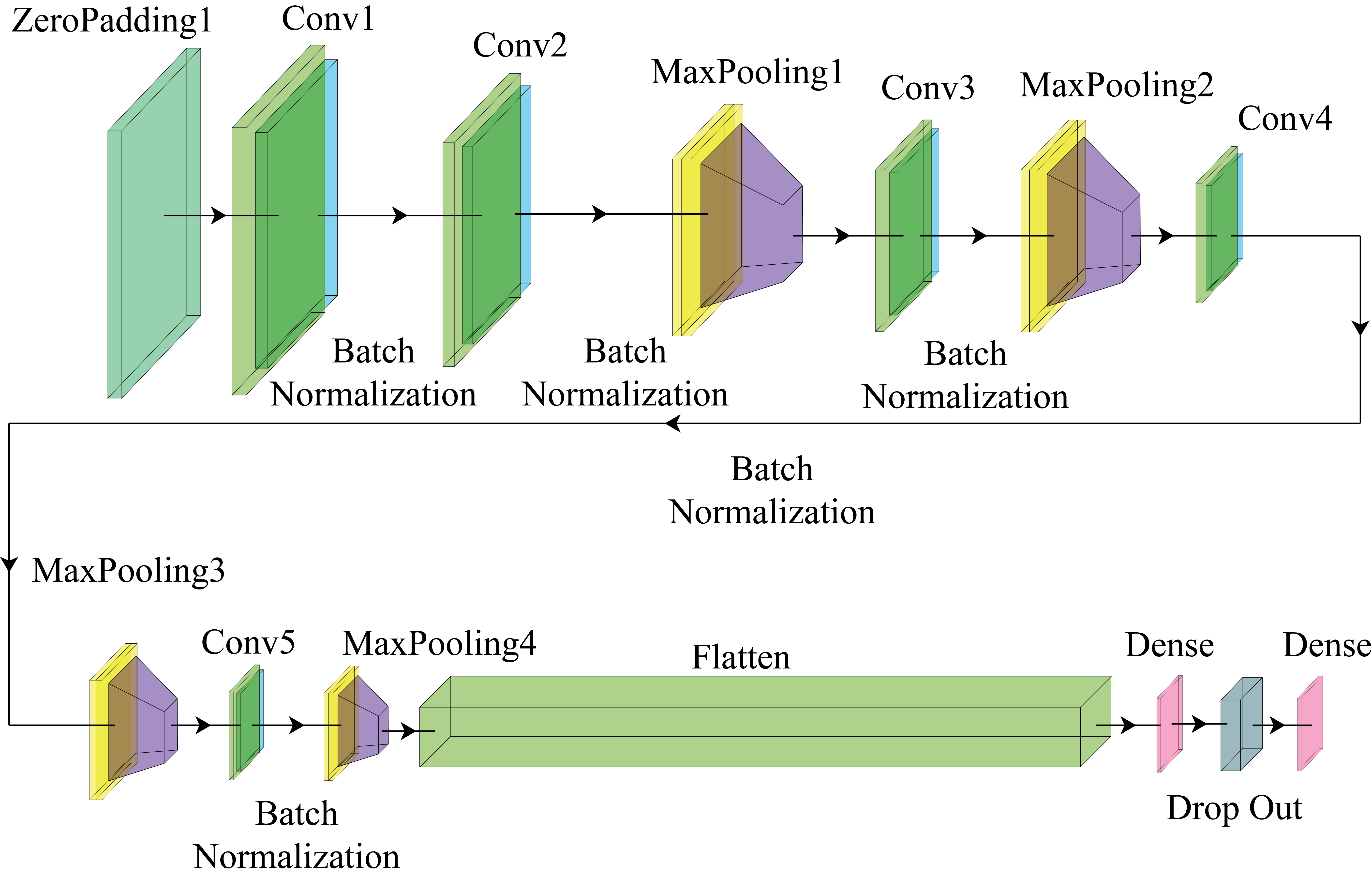}        
    }
    \caption{Illustration of the proposed deep CNN architecture}
    \label{architecture}
\end{figure*}

In this part, we present a Deep CNN-based architecture for performing the PPPC task. We begin by constructing a variation of the standard CNN framework with tweaks to key parameters and functions.
The structure of our network is depicted in Figure \ref{architecture}. 
It comprises a total of nineteen layers, which include one zero-padding layer, five convolutional layers, and two fully connected layers. The full model layers can be seen in Table \ref{parameter}.

Our DCNN model starts with a three-channel image obtained from the dataset, which is then resized to a resolution of $120\times200$ pixels. The ZeroPadding2D layer is first applied with a padding size of (3,3). Then, we employ five layers of Conv2D with a (3,3) kernel size. In order to decrease the computation cost, we use MaxPooling layers with a pool size of (2,2). Furthermore, the default strides for all Conv2D layers are (1,1).

We select the Rectified Linear Unit (ReLU) as our activation function because, in comparison to other activation functions like Tanh or Sigmoid, its unsaturated gradient significantly speeds up the stochastic gradient descent (SGD) computation. After that, a one-dimensional array made of the values is flattened. Furthermore, we extend the CNN architecture with 128 fully connected (FC) layers. In order to prevent overfitting, we insert a 10\% dropout rate between the Dense layer and the final output layer in this study. In the output layer, a classifier called the linear activation function is employed. A simple straight-line activation is produced by this linear activation function since it is directly proportional to the weighted sum of inputs or neurons. It allows for a broader range of activations, with the activation level increasing as the input rate rises along a favorably sloping line. 

Table \ref{hyperparameter} shows the optimum parameters applied in different datasets for the proposed DCNN model and Table \ref{parameter} shows the total parameter size in different layers of the proposed architecture.

\begin{table*}[!ht]
\centering
\resizebox{0.7\textwidth}{!}{%
\begin{tabular}{|cccccc|}
\hline
\rowcolor{lightgray}
\multicolumn{6}{|c|}{Hyperparameters} \\ \hline
\rowcolor{gray!20}
\multicolumn{1}{|c|}{Image Input Size} &
  \multicolumn{1}{c|}{Epoch} &
  \multicolumn{1}{c|}{Batch Size} &
  \multicolumn{1}{c|}{Learning Rate} &
  \multicolumn{1}{c|}{Dropout Rate} &
  Parameters \\ \hline
\multicolumn{1}{|c|}{120 x 200} &
  \multicolumn{1}{c|}{350} &
  \multicolumn{1}{c|}{8} &
  \multicolumn{1}{c|}{0.000001} &
  \multicolumn{1}{c|}{10\%} &
  4,849,601 \\ \hline
\end{tabular}

}
\caption{Hyperparameters of our proposed model}
\label{hyperparameter}
\end{table*}



\begin{table*}[!ht]
\centering
\resizebox{0.5\columnwidth}{!}{%
\begin{tabular}{|c|c|c|}
\hline
\rowcolor{lightgray}
\textbf{Layer}         & \textbf{Output Shape} & \textbf{No. of Parameters} \\ \hline \hline
Zero Padding 2D  & (None, 126, 206, 3)  & 0                 \\ \hline
Convolutional 2D  & (None, 124, 204, 32)  & 896                 \\ \hline
Batch Normalization    & (None, 124, 204, 32)  & 128                 \\ \hline
Convolutional 2D  & (None, 122, 202, 64)  & 18,496               \\ \hline
Batch Normalization    & (None, 122, 202, 64)  & 256                 \\ \hline
Max Pooling 2D          & (None, 61, 101, 64)    & 0                   \\ \hline
Convolutional 2D  & (None, 59, 99, 128)   & 73,856               \\ \hline
Batch Normalization    & (None, 59, 99, 128)   & 512                 \\ \hline
Max Pooling 2D          & (None, 29, 49, 128)   & 0                   \\ \hline
Convolutional 2D  & (None, 27, 47, 256)   & 2,95,168              \\ \hline
Batch Normalization    & (None, 27, 47, 256)   & 1,024                \\ \hline
Max Pooling 2D         & (None, 13, 23, 256)   & 0                   \\ \hline
Convolutional 2D  & (None, 11, 21, 512)   & 11,80,160             \\ \hline
Batch Normalization    & (None, 11, 21, 512)   & 2,048                \\ \hline
Max Pooling 2D         & (None, 5, 10, 512)    & 0                   \\ \hline
Flatten                & (None, 25600)         & 0                   \\ \hline
Dense                  & (None, 128)           & 32,896               \\ \hline
Dropout                & (None, 128)           & 0                   \\ \hline
Dense                  & (None, 1)             & 129                 \\ \hline
\end{tabular}
}
\caption{Layer-wise output shape and parameter size of the proposed model}
\label{parameter}
\end{table*}

\section{Experimental Evaluation}
\label{Experimental_Setup}

After constructing our proposed architecture, we evaluate the performance of our model using established metrics commonly used for regression-based tasks. We also compare the results of our model with those obtained from various existing deep learning architectures.
Since it is targeted to be implemented in mobile apps, our major goal is to facilitate a model that yields the optimal outcome, considering a lower need for computational resources.

\vspace{-0.1cm}

\subsection{Experimental Setup} \label{experiment}

All other popular CNN models, including our proposed Deep CNN model, are trained and tested using Tensorflow, Keras, Pillow, and OpenCV Python libraries. For the study, Python 3.6.13 is used alongside Tensorflow 2.6, Keras 2.6, and OpenCV 4.6. The models undergo training and evaluation on two distinct devices: one equipped with an NVIDIA RTX 2070 GPU delivering 7.5 TeraFLOPs of performance, and the other utilizing an NVIDIA RTX 3080TI GPU with a performance of 34.1 TeraFLOPs.

\subsection{Experimental Findings}


Four frequently utilized performance measures— Mean Absolute Error (MAE), Mean Square Error (MSE), Root Mean Square Error (RMSE), and R-squared ($R^2$) are taken into consideration when examining the prediction results to assess the performance of the proposed model. The metrics are shown in the following way:

\begin{equation}
M A E=\frac{1}{n} \sum_{t=1}^{n}\left|F_{p}-F_{t}\right|
\label{mae}
\end{equation}

\begin{equation}
M S E=\frac{1}{n} \sum_{t=1}^{n}\left(F_{p}-F_{t}\right)^{2}
\label{mse}
\end{equation}

\begin{equation}
R M S E=\sqrt{\frac{1}{n} \sum_{t=1}^{n}\left(F_{p}-F_{t}\right)^{2}}
\label{rmse}
\end{equation}

Here, $F_{p}$ represents the predicted value, while $F_{t}$ denotes the actual value.

\begin{equation}
R^{2}=1-\frac{\sum_{i=1}^{N}\left(y_{i}-y_{i}^{\prime}\right)^{2}}{\sum_{i=1}^{N}\left(y_{i}-\operatorname{avg}(y)\right)^{2}}
\end{equation}


$y_i$ refers to the forecasted value, $avg(y)$ represents the average forecasted value, and $y_i'$ signifies the $i^{th}$ observed value, where ${i=1, 2, .... N}$. R-squared is a metric that increases as the observed and forecasted values align, with a maximum value of 1 indicating a perfect match.








\begin{figure*}
\centering
\begin{subfigure}{.58\textwidth}
    \centering
    \includegraphics[width=1\linewidth]{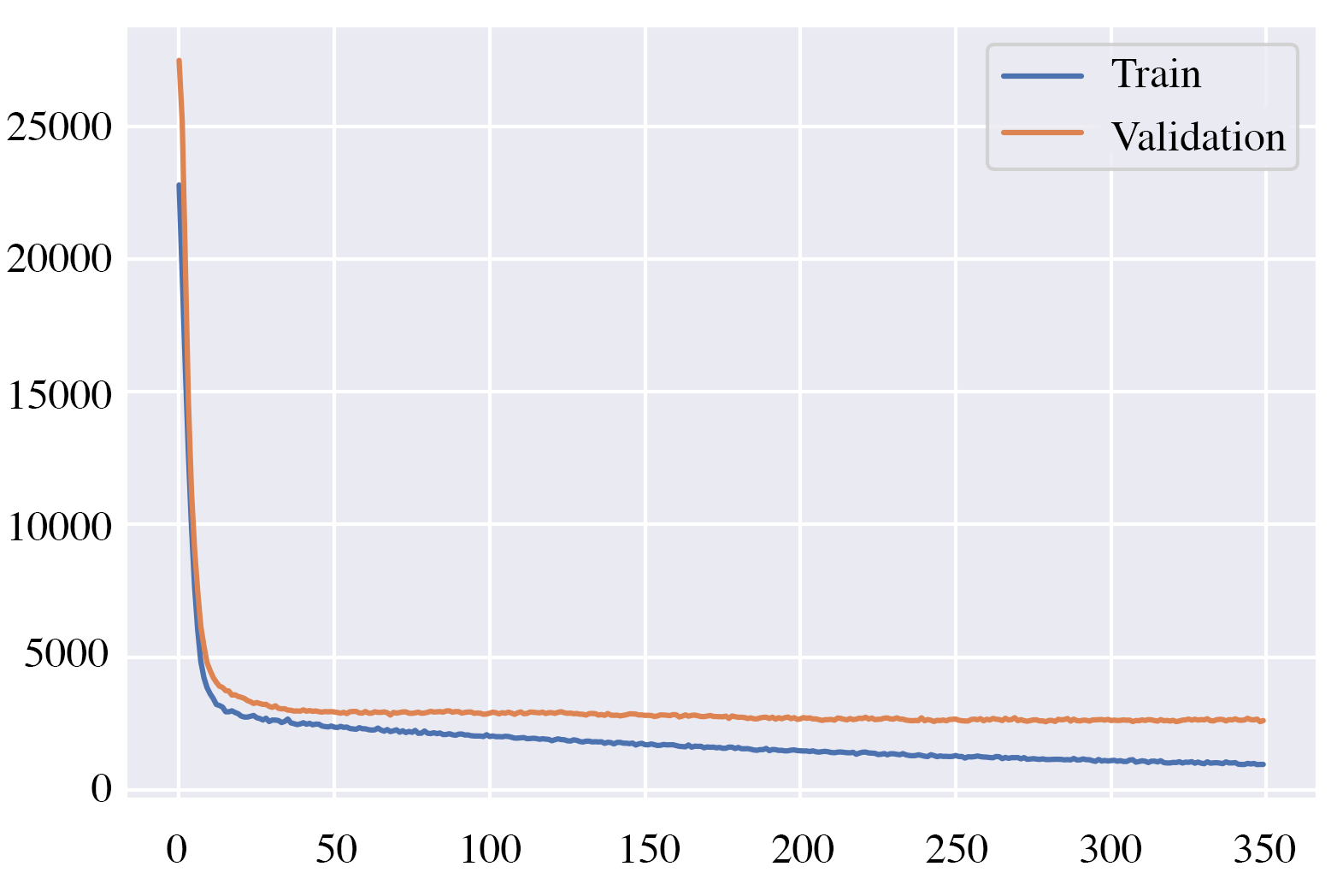}
    \caption{MSE}
    \label{mse-curve}
\end{subfigure}
\begin{subfigure}{.58\textwidth}
    \centering
    \includegraphics[width=1\linewidth]{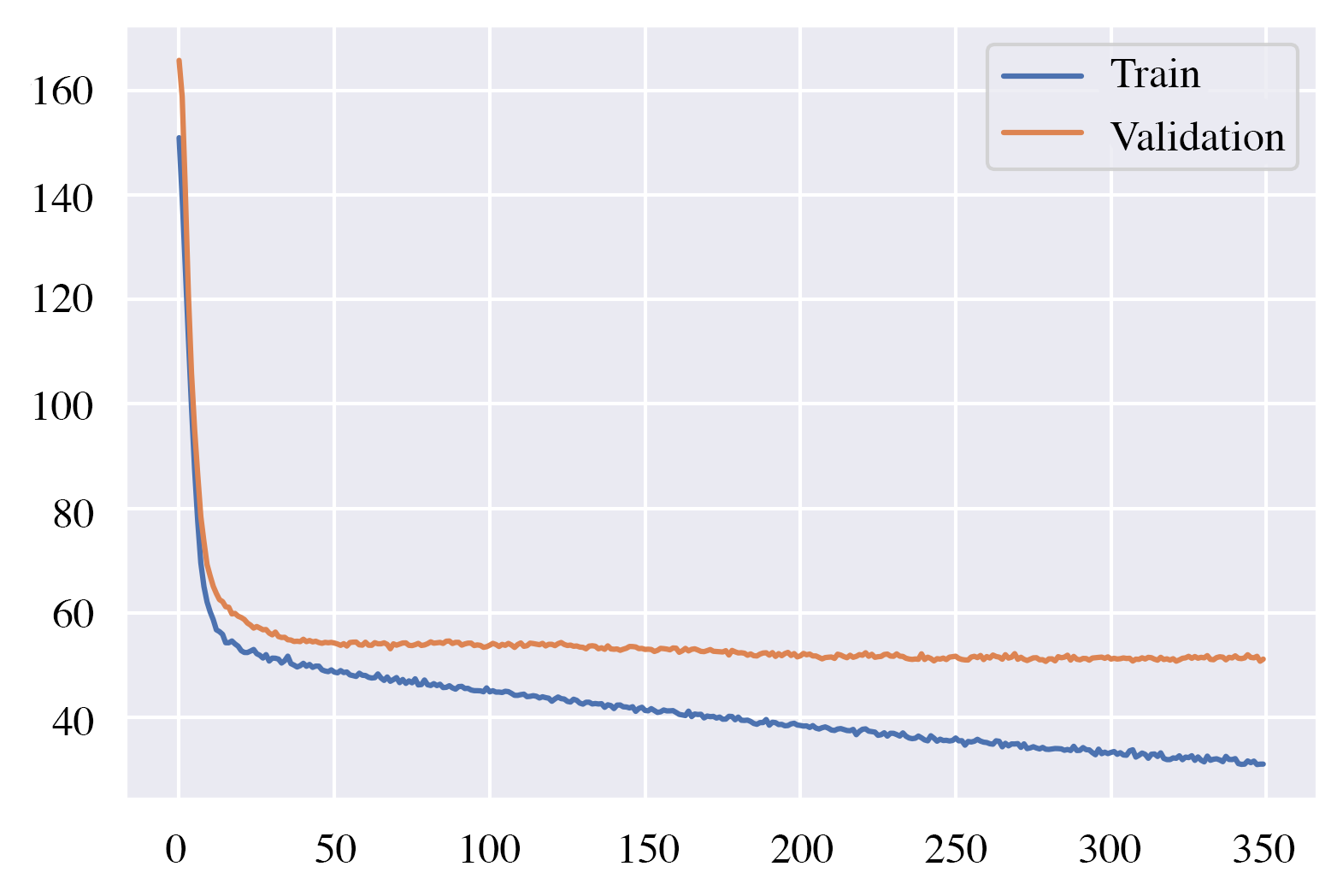}
    \caption{RMSE}
    \label{rmse-curve}
\end{subfigure}
\begin{subfigure}{.58\textwidth}
    \centering
    \includegraphics[width=1\linewidth]{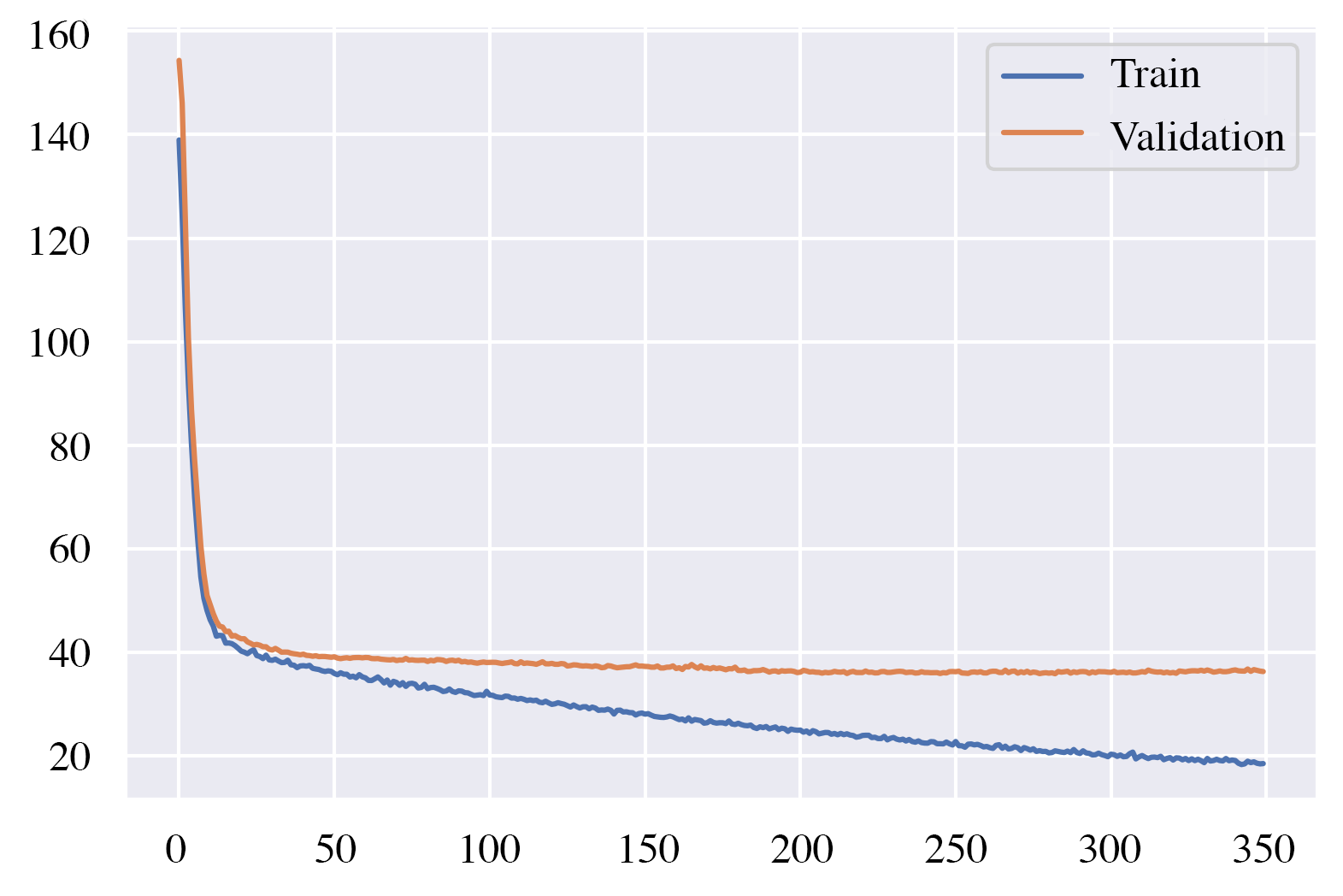}
    \caption{MAE}
    \label{mae-curve}
\end{subfigure}
\caption{Loss curves of the proposed model (the figures are created using Matplotlib library of Python version 3.6.13)}
\label{sample_data}
\end{figure*}

\begin{figure}[ht]

        \makebox[\linewidth]{

        \includegraphics[width=0.625\linewidth]{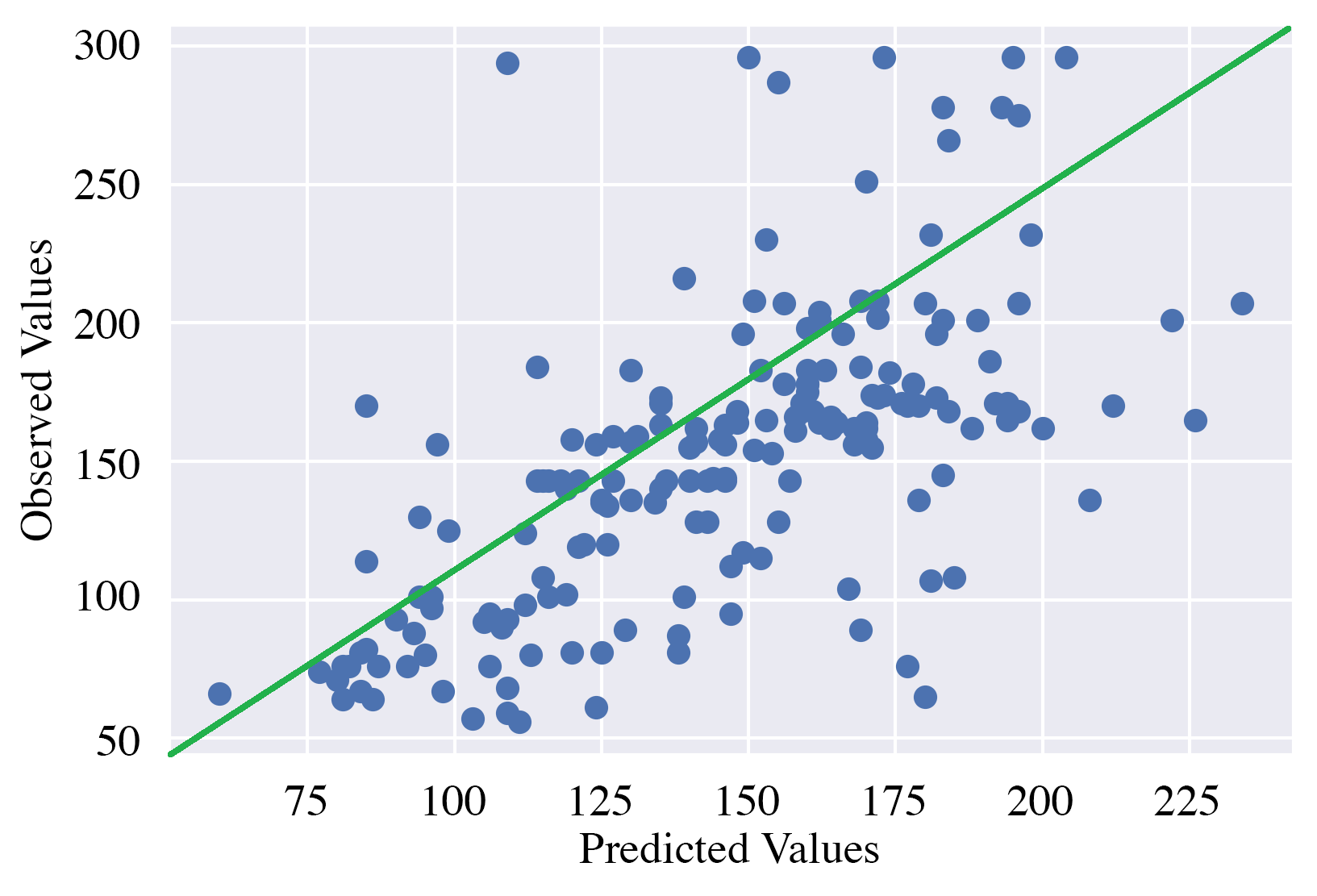}
    }
    

    \caption{Correlation between ground value/ observed PM2.5 index and estimated PM2.5 index [Here, the green line represents the expected predicted value against observed value.] (the figure is created using the Matplotlib library of Python version 3.6.13)}
    \label{pred-corr}
\end{figure}


\begin{table}
\centering
\resizebox{0.3\textwidth}{!}{%
\begin{tabular}{|c|c|c|c|}
\rowcolor{lightgray}
\textbf{MSE}     & \textbf{RMSE}  & \textbf{MAE}   & \textbf{$R^2$} \\
1782.08 & 42.22 & 29.68 & 0.387
\end{tabular}%
}
\caption{MSE, RMSE, MAE, and $R^2$ value for proposed model}
\label{results_final}
\end{table}

After investigating our study, we see an MSE of 1782.08, an RMSE of 42.22 an MAE of 29.68, and a $R^2$ of 38.7\% after running the model, as presented in Table \ref{results_final}. We also plot the MSE, RMSE, and the MAE of the train and validation data which are shown in Figure \ref{sample_data}. 
Our achieved $R^2$ value shows that around 39\% of the data fit with our regression model. If the predicted PM2.5 data and the observed PM2.5 data difference are below 50, it can be considered an acceptable prediction, as the PM2.5 value difference of below 50 is considered a close prediction. Moreover, we see the correlation coefficient of the model is 0.63, which is also an acceptable correlation in terms of real-world data. We present a scatter plot of predicted VS observed PM2.5 values based on the 182 test samples in Figure \ref{pred-corr}. The plot demonstrates the similarity of measurements. Our achieved value of correlation coefficient is more than 0.5 which means when one variable changes, the other variables change in the same direction. The most common interpretation of $R^2$ is how well the regression model fits the observed data.

To verify our model, we also run k-Fold Validation using our dataset where $k=10$. After running our model through the folds, we observe that the loss difference is less than 10\%. After evaluating our model, as no method in the literature can be compared directly for performance analysis with their reliance on different datasets, we compare it with some deep learning architectures available, including ResNet-50, VGG-19, InceptionV3, EfficientNetB2, and MobileNetV2 that use a pre-trained weight of ImageNet dataset. From Table \ref{testing-accuracy-table}, we can see that the majority of these methods utilize a high number of parameters overall, but our model employs a much lower percentage of parameters compared to others. Again, if we compare with MobileNetV2 which has a lesser amount of parameters than our approach, our approach shows significantly better results than MobileNetV2. 

In addition, we compare our model to prominent attention-based models such as Vision Transformer (ViT) and Involutional Neural Network (INN). The Transformer is mainly introduced by Vaswani et al. \cite{Vaswani2017-yz} and Dosovitskiy et al. \cite{vision_transformer_dosovitskiy} introduce the concept of Vision Transformer. They are image classification models built on the principles of Transformers. When an image is sent as input, the architecture divides that image into patches of specified size. Here, each patch is linearly embedded, positions are re-embedded, and the final vector sequence is sent to a conventional Transformer encoder. Adding an additional ``classification token'' to the sequence, which is learnable is the conventional method for doing classification.

Involutional Neural Network is primarily introduced by Li et al. \cite{Li2021-zu}. Involution is a process that inverts the design principles of convolution to be used in deep neural networks. Involution kernels are shared throughout channels even though they are spatially different. If involution kernels are specified as fixed-size matrices (similar to convolution kernels) and updated using the back-propagation process, the learned involution kernels cannot transfer across input pictures with different resolutions.

\begin{figure*}
\centering
\begin{subfigure}{.45\textwidth}
    \centering
    \includegraphics[width=1\linewidth]{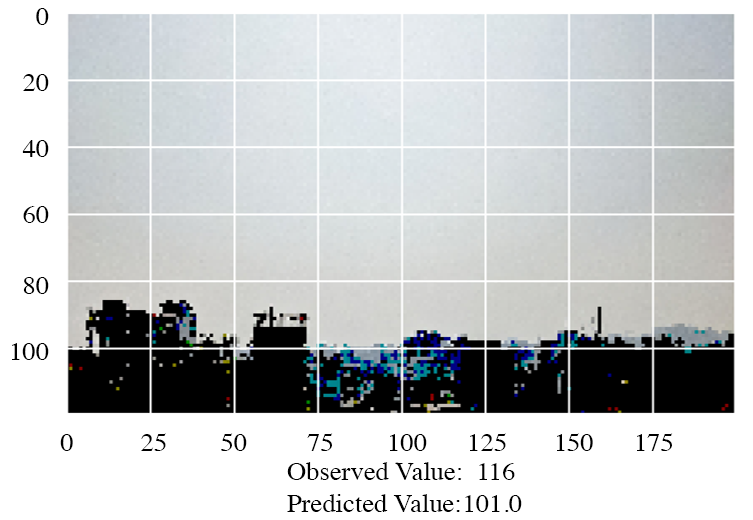}
    \caption{Sample Image 1}
    \label{1}
\end{subfigure}
\begin{subfigure}{.456\textwidth}
    \centering
    \includegraphics[width=1\linewidth]{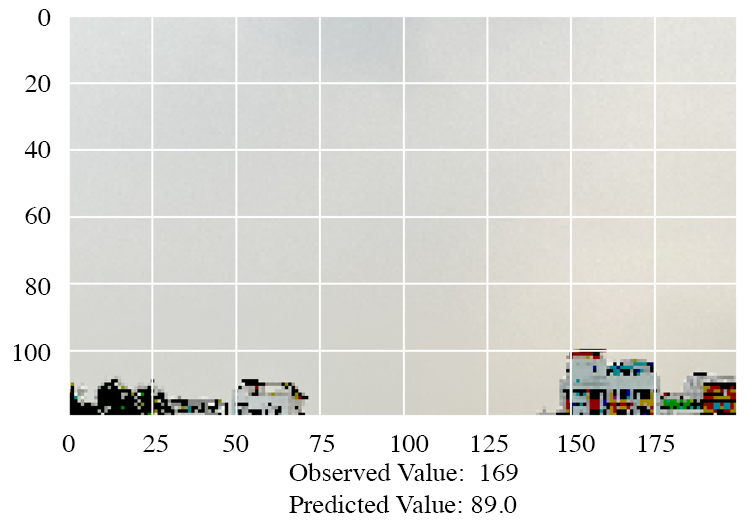}
    \caption{Sample Image 2}
    \label{2}
\end{subfigure}
\begin{subfigure}{.45\textwidth}
    \centering
    \includegraphics[width=1\linewidth]{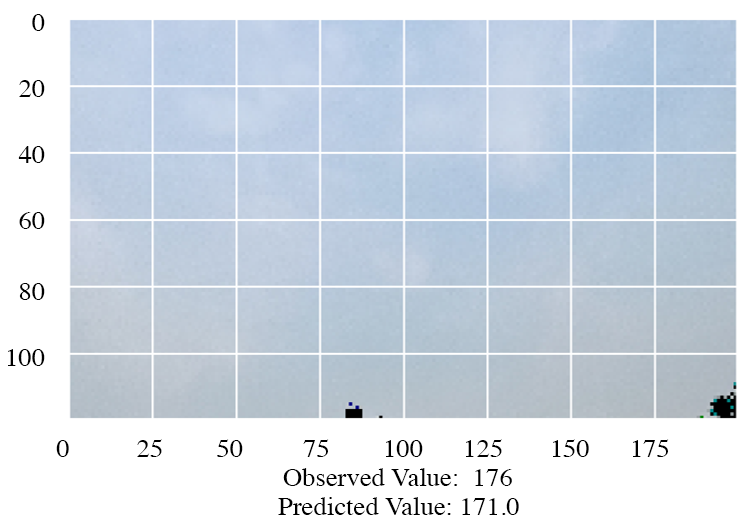}
    \caption{Sample Image 3}
    \label{3}
\end{subfigure}
\begin{subfigure}{.45\textwidth}
    \centering
    \includegraphics[width=1\linewidth]{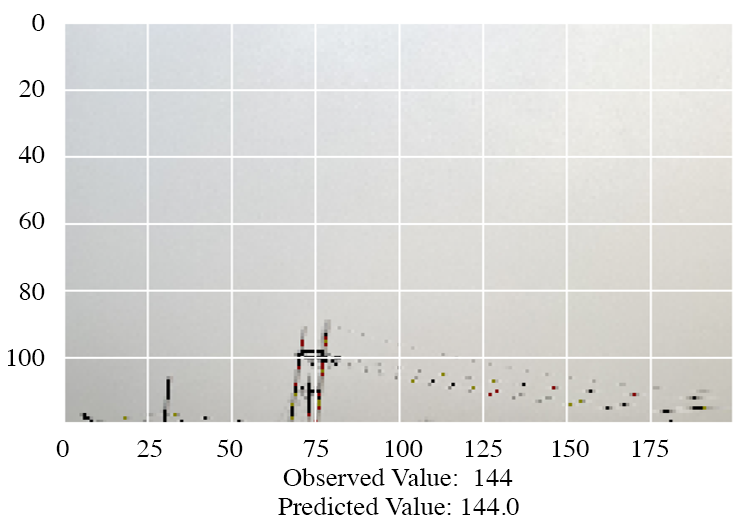}
    \caption{Sample Image 4}
    \label{4}
\end{subfigure}
\begin{subfigure}{.45\textwidth}
    \centering
    \includegraphics[width=1\linewidth]{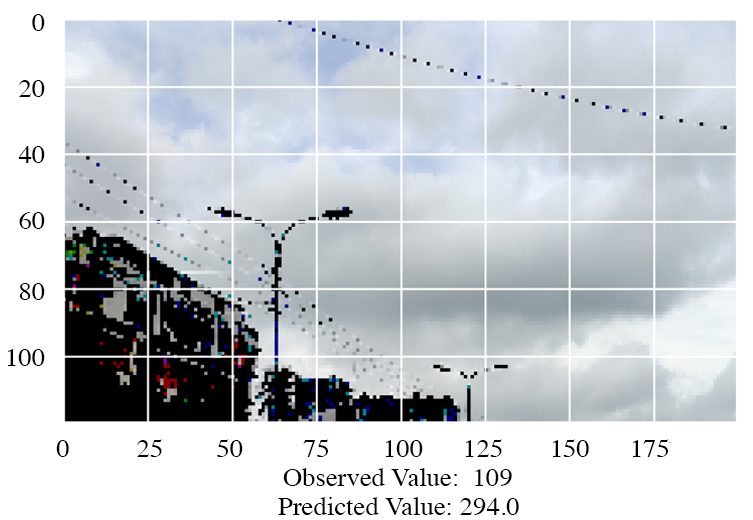}
    \caption{Sample Image 5}
    \label{5}
\end{subfigure}
\begin{subfigure}{.45\textwidth}
    \centering
    \includegraphics[width=1\linewidth]{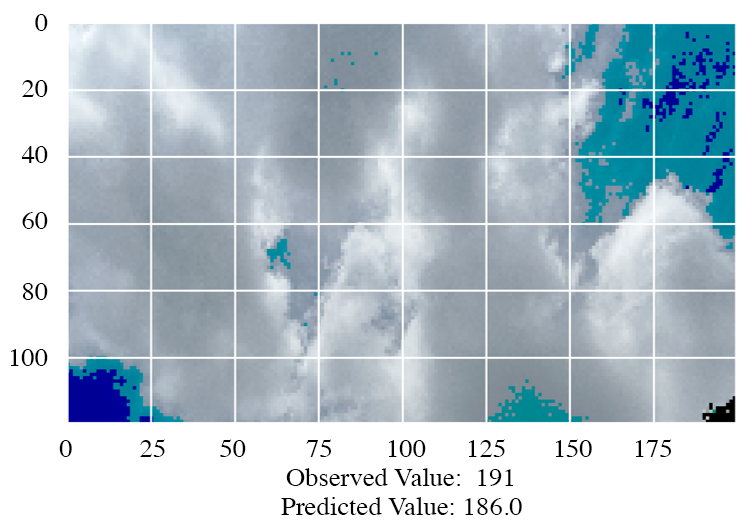}
    \caption{Sample Image 6}
    \label{6}
\end{subfigure}
\caption{Outputs on some sample test images of different days, on different areas [pre-processed] (the figures are created using the Matplotlib library of Python version 3.6.13, and these images are captured by one of the authors using smartphones)}
\label{output_images}
\end{figure*}

Besides, from Table \ref{testing-accuracy-table}, we can see that INN uses the lowest amount of parameters in total compared to our model but also performs the worst. If we compare with ViT which has an equivalent amount of parameters compared to other deep learning approaches, our approach shows significantly better results than ViT. Here, we can see negative $R^2$ values of all other models and it is usually negative when the chosen model does not follow the trend of the data. Moreover, from Table \ref{tab:comparison}, we can see that we outperform all of the existing deep learning-based approaches in terms of weight parameters. Figure \ref{output_images} shows some case scenarios where AQI from some of the images is almost correctly predicted and from the rest, the difference is really noticeably bad. 

\begin{table*}[ht]
\centering
\resizebox{0.8\textwidth}{!}{%
\begin{tabular}{|l|c|c|c|c|c|}
\hline
\rowcolor{lightgray}
\textbf{Deep Learning architectures} & \textbf{Parameters (in Millions)} & \textbf{MSE} & \textbf{RMSE} & \textbf{MAE} & \textbf{$R^2$} \\ \hline \hline
ResNet-50 & 23.6 & 2840.57 & 53.30 & 40.23 & 0.03 \\ \hline
VGG-19 & 20 & 8034.04 & 89.63 & 74.15 & -1.78 \\ \hline
InceptionV3 & 21.8 & 5243.73 & 72.41 & 55.16 & -0.80 \\ \hline
EfficientNetB2 & 64 & 2945.95 & 54.28 & 41.35 & -0.01 \\ \hline
MobileNetV2 & 2.2 & 2746.39 & 52.41 & 40.09 & 0.05 \\ \hline
ViT & 21.7 & 3304.33 & 57.48 & 44.66 & -0.14 \\ \hline
INN & \textbf{0.032} & 25397.24 & 159.37 & 149.93 & -7.79 \\ \hline
\textbf{Ours} & 4.8 & \textbf{1782.08} & \textbf{42.22} & \textbf{29.68} & \textbf{0.387} \\ \hline
\end{tabular}%
}
\caption{Comparison after Training Different Deep Learning Architectures in Our Dataset (Here, Epoch=350 for Each Model)}
\label{testing-accuracy-table}
\end{table*}

\begin{table}
\resizebox{\columnwidth}{!}{%
\begin{tabular}{|l|l|l|l|l|c|}
\hline
\rowcolor{lightgray}
Name & Underlying architecture & \begin{tabular}[c]{@{}l@{}}Number \\ of \\ parameters (Millions)\end{tabular} & Dataset & \begin{tabular}[c]{@{}l@{}}Classification \\ or Regression\end{tabular} & \begin{tabular}[c]{@{}l@{}}Image-based \\ test cases?\end{tabular} \\ \hline \hline
Liu et al. \cite{Liu2016-pc}, 2016 & \begin{tabular}[c]{@{}l@{}}Support vector regression \\ (SVR)\end{tabular} & N/A & Custom Datasets & Regression & \tikzxmark \\ \hline
Gu et al. \cite{Gu2019-mj}, 2019 & \begin{tabular}[c]{@{}l@{}}IL-NIQE \cite{Lin_Zhang2015-la}, ASIQE \cite{Gu2017-hl}, \\ LPSI \cite{Wu2015-cn}, NIQMC \cite{Gu2017-ly}, \\ BIQME \cite{Gu2018-tg}, S3 \cite{Vu2012-ow}, \\ FISH \cite{Vu2012-gz}, FISHbb \cite{Vu2012-gz}, \\ ARISMC \cite{Gu2015-hl} and BIBLE \cite{Li2016-nz}\end{tabular} & N/A & Custom Datasets & Regression & \tikzxmark \\ \hline
Zhang et al. \cite{ZHANG2020138178}, 2020 & \begin{tabular}[c]{@{}l@{}}SVM, \\ VGG-16, \\ ResNet18, \\ and Custom AQC-Net \\ (ResNet based CNN)\end{tabular} & \begin{tabular}[c]{@{}l@{}}N/A, \\ VGG-16: 138, \\ ResNet18: 11 \end{tabular} & Custom Datasets & Regression & \tikzxmark \\ \hline
Chakma et al. \cite{Chakma2017-ug}, 2017 & \begin{tabular}[c]{@{}l@{}}VGG-16 \\ and Random Forest\end{tabular} & VGG-16: 138, N/A & Custom Datasets & Classification & \checkmark \\ \hline
Li et al. \cite{Li_2015}, 2015 & \begin{tabular}[c]{@{}l@{}}Deep Convolutional \\ Neural Fields, \\ Dark Channel Prior\end{tabular} & N/A & FRIDA \cite{jpt-itsm12} & Classification & \checkmark \\ \hline
Rijal et al. \cite{Rijal2018-jt}, 2018 & \begin{tabular}[c]{@{}l@{}}VGG-16, \\ InceptionV3, \\ ResNet50\end{tabular} & \begin{tabular}[c]{@{}l@{}}VGG-16: 138, \\ InceptionV3: 25, \\ ResNet50: 23 \end{tabular} & \begin{tabular}[c]{@{}l@{}}Chakma et al. \cite{Chakma2017-ug} \\ + Extention\end{tabular} & Regression & \checkmark \\ \hline
Zhang et al. \cite{Zhang2016-kh}, 2016 & Custom CNN Model & N/A & Custom Dataset & Classification & \checkmark \\ \hline


Duong et al. \cite{Duong2020-cw}, 2020 & \begin{tabular}[c]{@{}l@{}}SVM, \\ Random Forest, \\ Extreme Gradient Boosting, \\ LightGBM and \\ CatBoost \cite{Prokhorenkova2017-mt}\end{tabular} & N/A & Custom Datasets & Mixed & \tikzxmark \\ \hline

Dao et al. \cite{Dao2021-te}, 2021 & \begin{tabular}[c]{@{}l@{}}FasterRCNN, \\ and Conditional-LSTM\end{tabular} & N/A & Custom Datasets\cite{visionair,Zhao2020-em,Nguyen-Tai2020-bn} & Mixed & \begin{tabular}[c]{@{}l@{}} \checkmark, but not \\ in all cases\end{tabular}   \\ \hline

Nilesh et al. \cite{Nilesh2022-ui}, 2022 & YOLOv5 & \begin{tabular}[c]{@{}l@{}}7.2 Million to \\ 86.7 Million\cite{Rath2022-wj}\end{tabular}   & Custom Dataset & Classification & \checkmark \\ \hline

Ahmed et al. \cite{Ahmed2022-eg}, 2022 & \begin{tabular}[c]{@{}l@{}}ResNet18-based \\ Custom Deep CNN\end{tabular} & N/A & Custom Dataset & Classification & \checkmark \\ \hline


Utomo et al. \cite{Utomo2023-rm}, 2023 & VGG-16 & 138 & Custom Dataset & Classification & \checkmark \\ \hline

Gu et al. \cite{Gu2022-zx}, 2022 & \begin{tabular}[c]{@{}l@{}}Custom DNN and \\ Nonlinear
Auto Regressive \\ Moving Average\end{tabular} & N/A & Custom Dataset & Regression & \tikzxmark \\ \hline

Wang et al. \cite{Wang2022-rj}, 2022 & Custom CNN-ILSTM & N/A & Custom Dataset & Regression & \tikzxmark \\ \hline

Janarthan et al. \cite{Janarthanan2021-jc}, 2021 & Custom SVR-LSTM & N/A & Custom Dataset & Regression & \tikzxmark \\ \hline

Our Model & Custom Deep CNN & 4 & Custom Dataset & Regression & \checkmark \\ \hline
\end{tabular}%
}
\caption{Comparison of our proposed approach with other existing PM2.5 classification/prediction based research studies (Here, N/A means either not mentioned or not applicable)}
\label{tab:comparison}
\end{table}


\section{Discussion}
\label{discussion}
In the preceding Sections, we present our results on PPPC utilizing deep learning methods. 
Our dataset is less skewed 
which is a crucial characteristic of a real-world dataset. Our achieved $R^2$ value is greater than 0.3 which is representative of real-world data. Also, our model, unfortunately, does not work well for noisy pictures. Moreover, it may not provide the expected result if the test image is taken outside Dhaka City since our training data is generated exclusively from the city. 
Since we have a lower amount of test data while running the correlation coefficient test, our correlation coefficient value also results lower than expected. 
When observing the resultant data, we see 81\% of the predicted values have 50 or fewer PM2.5 value differences. We keep 50 as a threshold for this because if the difference is less than 50 AQI, the impact on the environment can be considered similar.

As we work on other models, we first run them without any kind of pre-trained weights. After we run the models, we discover an intriguing feature of the situation. It has come to our attention that, apart from ResNet-50, every model that we evaluate is unable to even train correctly despite being given the same amount of epochs (350), which leads to inaccurate forecasts. Every model that is put through its test without weight pre-training predicts a value of 0 for every single outcome based on the test dataset.


Unlike larger models, our smaller model with five Conv2D blocks, Batch Normalization, and MaxPooling adequately extracts the features for solving our problem. In order to simplify the model and avoid overfitting, we add another dropout layer at the end with a 10\% dropout rate. Zero-padding is used to correlate to the data frame's time-limited assumption and more zero-padding results in the denser interpolation of the frequency samples around the unit circle. There is a common misconception that zero padding in the time domain results in greater spectral resolution in the frequency domain. In fact, we can construct networks more easily if we keep the height and width constant and pay less attention to the tensor dimensions when moving from one layer to another. Without padding, the size of the volume would shrink too rapidly. By keeping information at the edges, padding ends up enhancing performance. MaxPooling, in addition to preventing overfitting, reduces the computational cost by reducing the number of parameters to learn. Unlike other examined models that employ Average Pooling and choose smooth features, our model's usage of MaxPooling enables it to predict outcomes with more accuracy. We restrict the number of nodes in the Dense layer because employing a very deep fully-connected layer would be redundant since it would deliver the same performance but with more parameters.

While testing it on attention-based models, DCNN outperforms both ViT and INN. ViT and INN both have different learning approaches. ViT identifies patches and tries to find patterns from them, making it superior for tasks such as object detection or semantic segmentation. But in our problem, the whole image as a whole contains information, not a specific part of it, which makes ViTs \cite{vision_transformer_dosovitskiy} unsuitable. On the other hand, INNs are location-specific and channel-agnostic \cite{Li2021-zu}. While our task at hand is not specific to any color channels, the task of AQI recognition from images is spatially agnostic, since AQI is not computed from any certain location of the image, or should not be offset by the presence of any objects. There is no specific information in any particular segment of the images from which both ViT and INN look for information. This is why DCNN works better than both ViT and INN. 

Even though ViTs are reported to be relatively more resilient than DCNNs, DCNNs outperform ViT in this problem, as ViTs struggle when they have to extract any indistinct feature from an image. They are also more susceptible to contrast corruption as per 
Filipuk et al. \cite{filipiuk2022comparing}, which is a common source of noise in detecting AQI from images as human-made constructions appear in contrast to the surrounding ambiance. Moreover, Chen et al. \cite{https://doi.org/10.48550/arxiv.2112.09385} report that ViTs are more sensitive to noise and outliers. 




\subsection{Limitations}

When we work with the data, we notice that there are certain circumstances where the processing does not work well. For example, when the image has a direct reflection of the sun or when a bright sun is present in the picture, it then distorts the picture with excessive contrast which makes it harder for the model to predict. Besides, if the picture contains anything reflective of the sky, the model can not predict the PM2.5 accurately. In addition, due to the poor picture quality at night, visibility is often inadequate. This experimental model focuses only on daytime air quality monitoring (until evening) and cannot be declared acceptable for nighttime usage. Moreover, the data of each pixel carries significance toward the result. Therefore, lower-resolution cameras may provide some unexpected results since the pictures may contain noise. Noisy data in this case can heavily impact a negative outcome. 

In an effort to maintain fairness and commonality, we sought to optimize the hyperparameter models. In CNN-based models, we attempt to maintain it as optimal as the model originally receives, but optimal hyperparameters for each transformer model require more computational requirements. We keep the same hyperparameters as others for transformer models. This is a limitation of our study since the transformer-based models can be tuned and designed in numerous ways.

\section{Conclusion}
\label{Conclusion}
Assessing the PM2.5 index from an image is a challenging area of study since it necessitates comprehending the visible changes of multiple objects of varied depths, which demands a high degree of grasp of image contents. In this paper, we present a procedure for forecasting PM2.5 from pictures that use a DCNN model. The images are captured in such a way that at least 50\% of the sky is visible. The photos are then used as input, and the lower 50\% of the image is cropped to remove the part without the sky. The augmented image is then loaded into the suggested model. We learn via observations and analysis that the saturation of a picture might have a substantial influence on detecting the PM2.5 concentration from an image. So, based on that knowledge, we run CNN architecture in supervised mode, with the model receiving a picture and the AQI at the time span the image is captured. We test our custom DCNN on a custom dataset, and the findings show that CNN can be utilized to predict PM2.5 from photos and enhance estimation accuracy. We also execute a PM2.5 concentration picture collection, which presently consists of 1,818 photos with corresponding PM2.5 values, in addition to the suggested technique. 


Our future work plan includes converting the model to be used as a transfer learning model and implementing the model as a mobile app since our work totally focuses on being implemented in real life. Moreover, we will continue collecting more PM2.5 photos along with the PM2.5 of that exact location using AQI detection sensors from various contexts (e.g., cities, rural regions) from different time ranges
and different state-of-the-art models will be tested to get improved results. Additionally, we are concentrating on multimodal data collection, which combines images with information about temperature, humidity, wind speed, and date.

\section*{Data Availability}
All data generated or analyzed during this study will be made public when publishing the paper.

\bibliography{Bibliography}

\end{document}